\pdfoutput=1
\documentclass[runningheads]{llncs}
\usepackage{graphicx}

\usepackage{algorithm}
\usepackage{algpseudocode}

\usepackage{tikz}
\usepackage{comment}
\usepackage{amsmath,amssymb} 
\usepackage{color}
\usepackage{booktabs}
\usepackage{footnote}
\usepackage{caption,subcaption}
\usepackage{multirow}
\usepackage{color}
\usepackage[accsupp]{axessibility}  
\usepackage[bottom]{footmisc}

\begin{document}

\pagestyle{headings}
\mainmatter
\def\ECCVSubNumber{2031}  

\title{Efficient Meta-Tuning for Content-aware Neural Video Delivery} 

\titlerunning{Efficient Meta-Tuning(EMT)}
%
\author{Xiaoqi Li\inst{1}\thanks{Equal contribution}\and Jiaming Liu\inst{1,2*} \and Shizun Wang\inst{3*} \and Cheng Lyu\inst{3} \and Ming Lu\inst{4\dag} \and Yurong Chen\inst{4} \and Anbang Yao\inst{4} \and Yandong Guo\inst{2}\and Shanghang Zhang\inst{1}\thanks{Corresponding Author}}
\authorrunning{X. Li et al.}
%
\institute{Peking University \and
OPPO Research Institute \and
Beijing University of Posts and Telecommunications \and
Intel Labs China \\
\email{clorisleef0313@gmail.com, shzhang.pku@gmail.com, yandong.guo@live.com}}

\maketitle

\begin{abstract}
Recently, Deep Neural Networks (DNNs) are utilized to reduce the bandwidth and improve the quality of Internet video delivery. Existing methods train corresponding content-aware super-resolution (SR) model for each video chunk on the server, and stream low-resolution (LR) video chunks along with SR models to the client. Although they achieve promising results, the huge computational cost of network training limits their practical applications. In this paper, we present a method named Efficient Meta-Tuning (EMT) to reduce the computational cost. Instead of training from scratch, EMT adapts a meta-learned model to the first chunk of the input video. As for the following chunks, it fine-tunes the partial parameters selected by gradient masking of previous adapted model. In order to achieve further speedup for EMT, we propose a novel sampling strategy to extract the most challenging patches from video frames. The proposed strategy is highly efficient and brings negligible additional cost. Our method significantly reduces the computational cost and achieves even better performance, paving the way for applying neural video delivery techniques to practical applications. We conduct extensive experiments based on various efficient SR architectures, including ESPCN, SRCNN, FSRCNN and EDSR-1, demonstrating the generalization ability of our work. The code is released at \url{https://github.com/Neural-video-delivery/EMT-Pytorch-ECCV2022}.
\end{abstract}

\section{Introduction}
\label{sec:intro}

With the popularity of High-Definition (HD) display devices, high-resolution videos are strongly demanded by end users. This brings a huge burden to the video delivery infrastructure. As the development of deep learning, several recent works are proposed to reduce the bandwidth of video delivery \cite{kim2020neural,yeo2018neural,liu2021overfitting,khani2021efficient}. The motivation of these works is to stream both the low-resolution videos and content-aware SR models from servers to clients. The clients run the inference of SR models to super-resolve the LR videos. In this manner, high-resolution videos can be delivered under limited Internet bandwidth.

In contrast to existing approaches on Single Image Super-Resolution (SISR) \cite{shi2016real,dong2014learning,lim2017enhanced,zhang2018image,kim2016accurate} and Video Super-Resolution (VSR) \cite{caballero2017real,wang2019edvr,chan2020basicvsr,chan2021basicvsr++}, content-aware models utilize the overfitting property of DNNs to achieve higher SR performance. To be more specific, a video is divided into several video chunks, and a corresponding SR model is trained for each chunk. This type of DNN-based video delivery system can achieve better performance even compared with commercial techniques like WebRTC \cite{kim2020neural}. 

Although neural video delivery is a promising technique, the huge computational cost of training content-aware SR models limits its practical applications. For example, existing methods \cite{liu2021overfitting,yeo2018neural} uniformly divide a 45s/1080P/30FPS video into 5-second chunks, and train the SR models for all chunks. However, even with efficient SR architectures like ESPCN \cite{shi2016real}, it still takes about 10.2 hours to train the content-aware SR models on a high-end NVIDIA V100 GPU. Therefore, reducing the computational cost of network training is crucial for neural video delivery. 

In order to pave the way for practical applications, we propose Efficient Meta-Tuning (EMT) in this paper. Instead of training from scratch \cite{liu2021overfitting,yeo2018neural}, EMT sequentially adapts a meta-learned model to the video chunks, delivering all the content-aware SR models. Compared with random initialization or pre-trained initialization, a meta-learned model can transfer better to different video chunks. We collect a large-scale dataset of diverse video chunks and take each chunk as one specific task. MAML \cite{finn2017model} is adopted to train the meta-learned model, whose parameters are shared by all content-aware SR models. For the chunks of the input video, EMT adapts the meta-learned model to the first chunk. As for the following chunks, it can fine-tune the partial parameters of the previous adapted model due to the temporal consistency between neighboring chunks. The partial parameters are selected by gradient masking, which masks a fraction of most significant parameters after a few gradient updates. Since EMT sequentially adapts the meta-learned model, each chunk simply needs to store the selected partial parameters. The current content-aware SR model can be constructed by updating the partial parameters of the previous model. This is important to compress all the models into one shared model and a few private parameters. Compared with CaFM \cite{liu2021overfitting}, our method is more compact since the meta-learned model is shared by all chunks, while CaFM can only share one model for chunks within the input video.

To further reduce the computational cost, we propose a novel sampling strategy for EMT, which selects the most challenging patches from video frames. Our motivation is that previous adapted SR model already possesses the ability to super-resolve current chunk due to temporal consistency. Therefore, the training efforts of EMT should focus on challenging patches, which cannot be well handled by the previous model. However, performing the evaluation of previous model on all patches of current chunk is time-consuming and brings additional cost. Inspired by video codec, we extract the I-frames from the input video and only perform the evaluation on I-frames. The positions of challenging patches are extracted based on I-frames and propagated to other frames. Since I-frame is very sparse within a video, the computational cost of the evaluation is negligible. On the other side, as the frames between two I-frames are temporally consistent, the propagated positions can extract reasonable patches on the in-between frames. Our sampling strategy is simple yet effective and can further reduce the computational cost of EMT.

Our contributions can be concluded as follows:
\begin{itemize}
\item We propose Efficient Meta-Tuning (EMT) for neural video delivery, significantly reducing the cost of training content-aware SR models and achieving even better performance.

\item We present a novel challenging patch sampling strategy, which further reduces the cost of EMT. Our strategy improves the convergence of EMT with negligible additional cost.

\item We conduct detailed experiments based on various efficient SR architectures to evaluate the advantage and generalization of our method.
\end{itemize}

\section{Related work}

{\bf DNN-based Image Super-Resolution} SRCNN \cite{dong2014learning} is the first work that introduces DNNs to SR task. Their method consists of three stages, namely feature extraction, non-linear mapping and image reconstruction. With the rapid advance of DNN, plenty of methods are proposed to improve the performance of SISR following the pipeline of SRCNN. For example, VDSR \cite{kim2016accurate} adopts a very deep DNN to predict the image residual instead of HR image. Motivated by ResNet \cite{he2016deep}, SRResNet \cite{ledig2017photo} introduces Residual Block to the network and improves the SR performance. EDSR \cite{lim2017enhanced} modifies the structure of SRResNet by removing the Batch Normalization layer \cite{ioffe2015batch}, further boosting the SR performance. RCAN \cite{zhang2018image} introduces the attention mechanism to the networks and presents deeper DNNs for SR. However, RCAN is computationally complicated, which limits its practical usage. To reduce the computational cost, many efficient methods are proposed for SR. ESPCN \cite{shi2016real} uses LR image as input and up-samples the feature map by the pixel-shuffle layer to obtain the HR output. LAPAR \cite{li2020lapar} proposes a method based on linearly-assembled adaptive regression network. All of those methods are external methods, which train one model on large-scale image databases like DIV2K \cite{Agustsson_2017_CVPR_Workshops} and test on given input images. However, external methods fail to explore the overfitting property of DNNs, which can significantly boost the performance for practical video delivery system. 

{\bf DNN-based Video Super-Resolution} Different from image super-resolution, video super-resolution can additionally exploit the neighboring frames for SR. Therefore, temporal alignment plays an essential role and should be thoroughly considered. VESPCN \cite{caballero2017real} first predicts the motions between neighboring frames, and then performs image warping before feeding neighboring frames into the SR network. However, it is difficult to accurately estimate the optical flow. TOFlow \cite{xue2019video} proposes a task-oriented flow designed for specific video processing tasks. They jointly train the motion estimation component and video processing component in a self-supervised manner. DUF \cite{jo2018deep} solves the problem of accurate explicit motion compensation by training a network to generate dynamic up-sampling filters and a residual image. In order to reduce the computational cost of VSR, FRVSR \cite{sajjadi2018frame} presents a recurrent framework that uses the previous SR result to super-resolve the following frame. Their recurrent framework naturally ensures temporally consistency and reduces the computational cost by warping only one image in each step. All these VSR approaches also belong to external methods that fail to explore the overfitting property of DNN. Apart from this, handling temporal alignment brings huge additional computational and storage costs, which limits their practical applications in resource-limited devices like mobile phone.

{\bf Neural Video Delivery} NAS \cite{yeo2018neural} is a promising Internet video delivery framework that integrates DNN for quality enhancement. It can solve the video quality degradation problem under limited Internet bandwidth. NAS can enhance the average Quality of Experience (QoE) by $43.08\%$ using the same bandwidth budget, or saving $17.13\%$ of bandwidth while providing the same user QoE. The main idea is to leverage DNN's overfitting property and use the training accuracy to deliver high SR performance. Many following works are proposed to apply the idea of NAS to different scenarios, like UAV video streaming \cite{xiao2019sensor}, live streaming \cite{kim2020neural}, 360 video streaming \cite{dasari2020streaming,chen2020sr360}, volumetric video streaming \cite{zhang2020mobile}, and mobile video streaming \cite{yeo2020nemo}, etc. Recent methods \cite{liu2021overfitting,khani2021efficient} propose to further reduce the bandwidth budget by sharing most of the parameters over video chunks. Therefore, only a small portion of private parameters are streamed for each video chunk. However, they still need huge computational cost for network training and fail to study the scene conversion for constructing optimal video chunks.

\section{Method}
\subsection{Overview}
\begin{figure*}
\begin{center}
	\includegraphics[width=12cm, height=5cm]{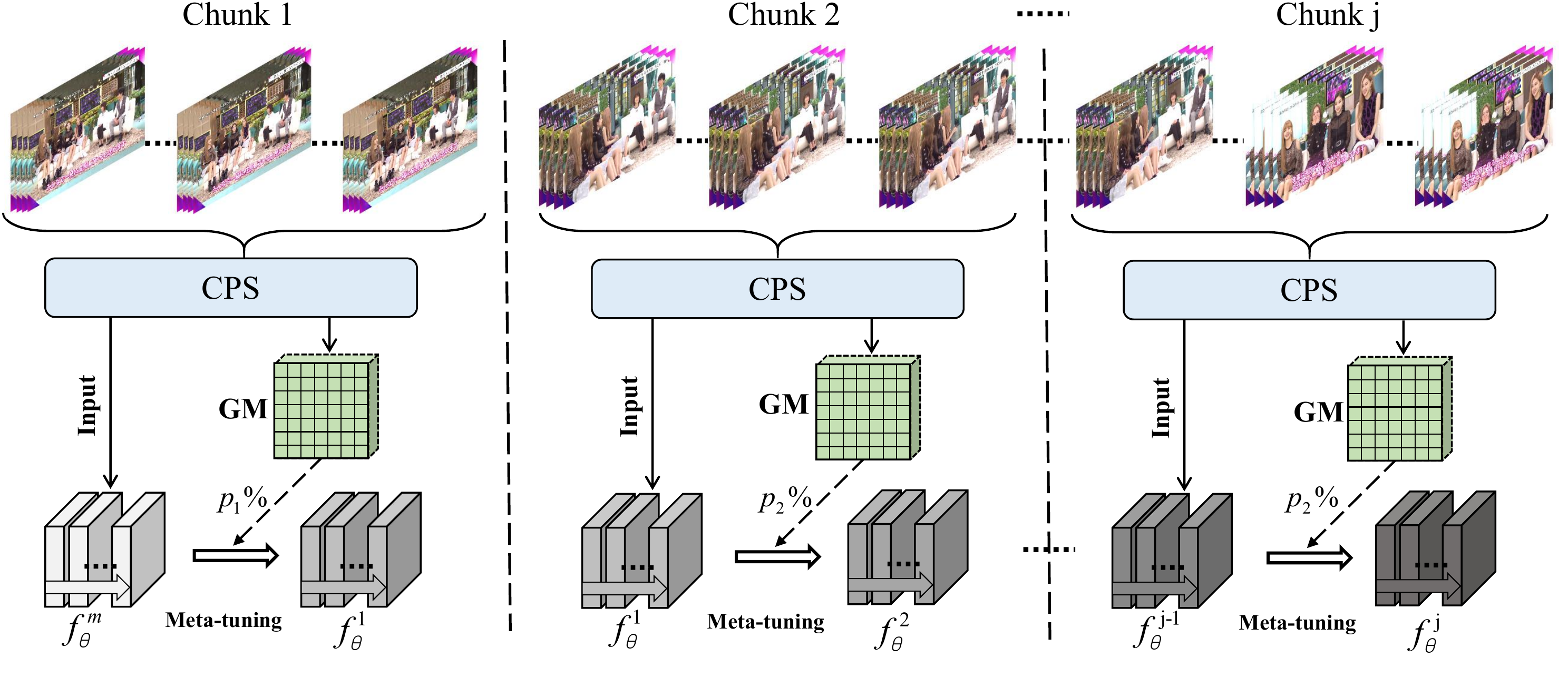}
\end{center}
\caption{The pipeline of our method. CPS and GM indicate challenging patch sampling and gradient masking respectively. The selected challenging patches are used to mask the parameters and fine-tune the models.}
\label{fig:pipeline}
\end{figure*}

In this section, we present our method to significantly accelerate the training of content-aware SR models. Our method adapts the meta-learned model to the first chunk of the video, and sequentially adapts partial parameters of the previous model to the following chunks. The partial parameters are selected by gradient masking and the challenging patches are extracted for adaption. The pipeline of our method is illustrated in Fig. \ref{fig:pipeline}. We first introduce Efficient Meta-Tuning (EMT) to sequentially deliver the models from a meta-learned model in Sec. \ref{sec:emt}. Then we propose a novel challenging patch sampling strategy to further accelerate EMT in Sec. \ref{sec:cps}. 
 
\subsection{Efficient Meta-Tuning}
\label{sec:emt}

Following former works \cite{liu2021overfitting,yeo2018neural}, we uniformly divide the input video into chunks, and train the corresponding content-aware SR model for each chunk. \cite{yeo2018neural} proposes to apply deep super-resolution networks to video delivery by training one model for each chunk from scratch. \cite{liu2021overfitting} presents a method to compress all the models by one shared model and a few private parameters. However, both methods train the content-aware SR models from scratch, resulting in huge computational cost. Since neighboring chunks are temporally consistent, fine-tuning is much more reasonable compared with training from scratch. Precisely, finding a generic initial model that can not only generalize over diverse video chunks but also adapt rapidly to any specific video chunk, plays a key role in fine-tuning. In order to obtain a better initialization, we adopt MAML \cite{finn2017model} to train a meta-learned model. Although MAML has been applied to Zero-Shot SR \cite{park2020fast,soh2020meta} and video frame interpolation \cite{choi2020scene}, it has never been studied in neural video delivery to the best of our knowledge. Compared with random initialization or pretrained initialization, a meta-learned model has better transferability. To obtain the content-aware SR models, we sequentially fine-tune partial parameters of the previous model. In contrast to fine-tuning the whole model, our method can compress the parameters of all models into one shared meta-learned model and a few partial parameters.

{\bf Meta-Learned Initialization} We take one chunk as a specific task and aim to learn a SR model that can adapt to various chunks. Specifically, we first pretrain the SR model $f_{\theta}$ on DIV2K \cite{Agustsson_2017_CVPR_Workshops}, then we utilize meta learning to optimize the pretrained parameters as illustrated by Alg. \ref{alg:maml}. This step enables the SR model to converge to a transferable point, which can be rapidly fine-tuned. In order to build a variety of tasks, we collect several video sequences and uniformly divide them into video chunks. Totally, we obtain $N$ chunks for meta-learning, and each chunk is set as the task $t_{i}$. The collected dataset is denoted as $D_{N}$. We apply bicubic downsampling to the frames and generate LR-HR pairs $(I_{LR}^{i},I_{HR}^{i})$. Our goal is to optimize $f_{\theta}$ according to each LR-HR pair by minimizing the L1 loss as shown in Eq. \ref{eq:1}. 

\begin{equation}
\label{eq:1}
\mathcal{L}_{f} = |f_{\theta}(I_{LR})-I_{HR}|_{1}
\end{equation}

During the inner loop (Line 4-7), we conduct one or more gradient updates for the task $t_i$ in each iteration. The temporary model for task $t_{i}$ is denoted as $f_{\theta i}$. During each inner gradient update, the task-specific parameters are updated according to Eq. \ref{eq:2}, where $\alpha$ is the inner learning rate.

\begin{equation}
\label{eq:2}
f_{\theta i}\gets f_{\theta i}-\alpha \nabla_{f_{\theta i}} \mathcal{L}_{f_{\theta i}}
\end{equation}

As for the outer loop (Line 9-10), we evaluate the loss of $f_{\theta i}$ on each $t_i$ and sum up the losses of all tasks to update the SR model $f_{\theta}$. For one outer gradient update, it considers the gradients from all tasks. The outer update can be formulated as Eq. \ref{eq:3}, where $\beta$ is the outer learning rate.

\begin{equation}
\label{eq:3}
f_{\theta}\gets f_{\theta}-\beta \nabla_{f_{\theta}}  \sum_{i} \mathcal{L}_{f_{\theta i}}
\end{equation}

\begin{algorithm}[t]
	\caption{Meta-Learned Initialization}
	\label{alg:maml}
	\hspace*{0.02in} {\bf Input:}
	Initialized SR model $f_{\theta}$, meta-learning dataset $D_{N}$ \\
	\hspace*{0.02in} {\bf Output:} 
	Meta-learned model $f_{\theta}^{m}$
	\begin{algorithmic}[1]
		\While{not done}
		\State Sample n tasks ${D_n}$ from $D_{N}$
		\For{${t_i} \in {D_n}$} 
		\State Sample pairs $(I_{HR}^{i}, I_{LR}^{i})$ from ${t_i}$
		\State Copy $f_{\theta i}$ from the latest $f_{\theta}$
		\State Evaluate training loss according to Eq. \ref{eq:1}
		\State Update parameters according to Eq. \ref{eq:2}
		\EndFor
		\State Calculate $\mathcal{L}_{f_{\theta i}}$ with respect to ${t_i}$
		\State Update $f_{\theta}$ according to Eq. \ref{eq:3}
		\EndWhile
		\State \Return Meta-learned model $f_{\theta}^{m}$
	\end{algorithmic}
\end{algorithm}

{\bf Partial Model Adaption} The meta-learned model $f_{\theta}^{m}$ is shared by all the video chunks. To obtain the content-aware SR models for the input video, we adapt the meta-learned model to the first chunk, and sequentially fine-tune the partial parameters of previous adapted model. Formally, we denote the content-aware SR model of ${j^{th}}$ chunk as $f_{\theta}^{j}$. For the first chunk, we rapidly fine-tune the meta-learned model $f_\theta^{m}$ to obtain the adapted model $f_{\theta}^{1}$. We use the gradient masking to select ${p_1}\%$ most significant parameters before adapting $f_{\theta}^{m}$. As for other chunks, we fine-tune previous model $f_{\theta}^{j-1}$ on ${j^{th}}$ chunk, delivering the adapted model $f_{\theta}^{j}$. We also adopt the gradient masking to select ${p_2}\%$ most significant parameters before fine-tuning $f_{\theta}^{j-1}$. It has to be noted that we adopt different percentages for the first chunk and other chunks in this work. Our partial model adaption requires much fewer epochs for both the first chunk and other chunks. Therefore, the computational cost can be greatly reduced compared with training from scratch under the same performance.

{\bf Gradient Masking} In order to compress the parameters of content-aware models, we design a simple yet effective strategy to find the ${p}\%$ most significant parameters. Given a reference model $f$, we need to find a fraction of parameters before fine-tuning $f$. Specifically, we adopt a few iterations to update $f$, obtaining a temporal model $\widehat f$. Afterwards, we calculate $|{\theta _{\widehat f}}-{\theta _f}|$ and choose the ${p}\%$ parameters that vary most, delivering the parameter mask $M({\theta _f})$.


Once we collect the $p\%$ most significant parameters, we can fine-tune the reference model $f$ and simply update the significant parameters. In this manner, our method only needs to store $p\%$ private parameters of the reference model. When fine-tuning for the first chunk, we choose the meta-learned model $f_{\theta}^{m}$ as the reference model. For ${j^{th}}$ chunk of the input video, we choose the previous adapted model $f_{\theta}^{j-1}$ as the reference model.

\subsection{Challenging Patch Sampling}
\label{sec:cps}

\begin{figure*}[t]
	\begin{center}
		\includegraphics[width=12cm, height=5cm]{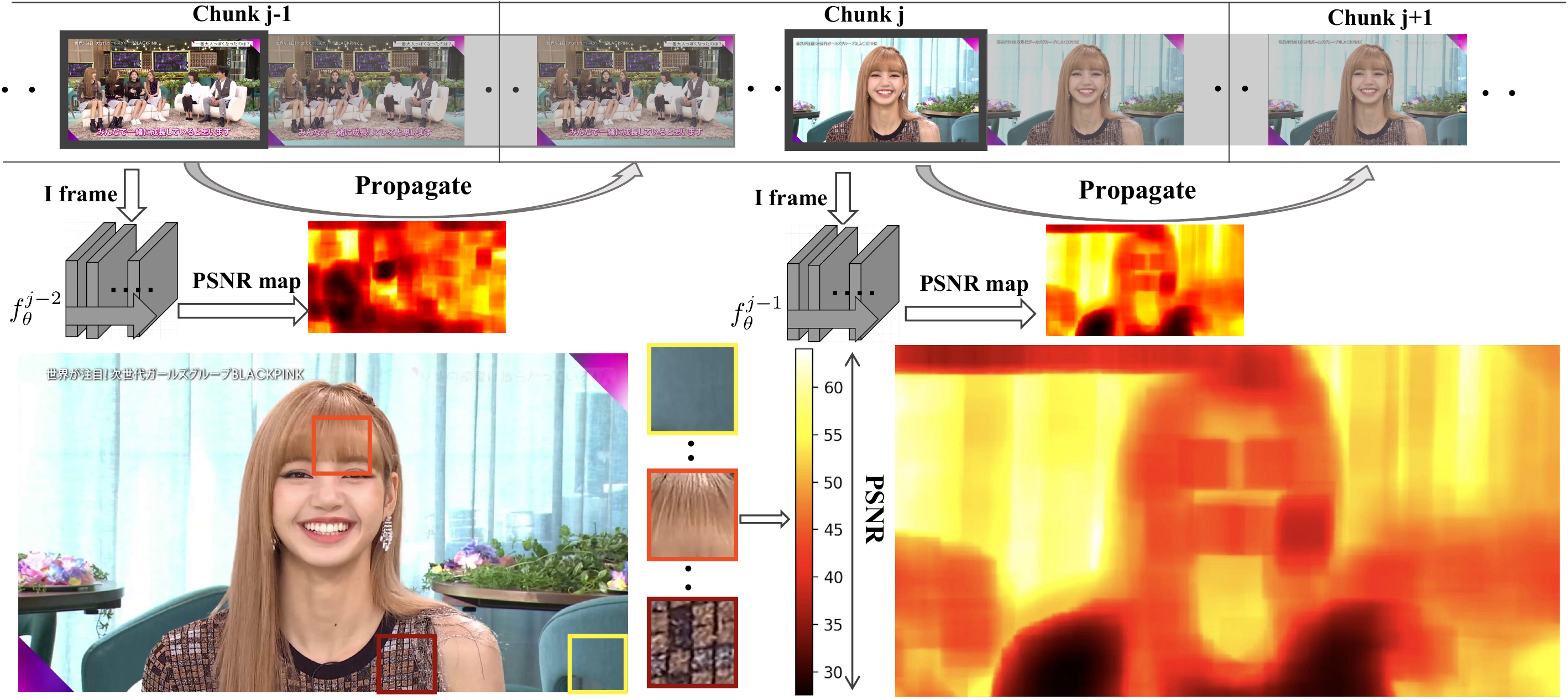}
	\end{center}
	\caption{Illustration of challenging patch sampling. Different colors in PSNR map represent patch's difficult levels for previous chunk's model. 'propagate' arrow indicates the propagation of PSNR map from I frame to its in-between frames.}
	\label{fig:cps}
\end{figure*}

Previous adapted SR model already possesses the ability to super-resolve current chunk due to the temporal consistency. Thus it can achieve satisfying results on the majority of regions. However, some hard regions are still challenging for the previous adapted model. Since SR networks are fine-tuned on sampled LR-HR patch pairs, we further present a strategy to sample the ${r\%}$ most challenging patches for EMT. Our strategy is highly efficient and brings negligible additional cost. Inspired by the video codec, it first locates the positions of challenging patches in I-frames, then propagates the positions to in-between frames as shown in Fig. \ref{fig:cps}.

Formally, we denote the frames of input video as ${I_1},...,{I_T}$, where $T$ is the number of frames. The I-frames are denoted as ${I_{{k_1}}},...,{I_{{k_M}}}$, where ${k_1},...,{k_M}$ are the indices of $M$ I-frames. We also denote the chunk indices of $M$ I-frames as ${c_1},...,{c_M}$. In order to localize the challenging patches for ${I_{{k_m}}}$, we run the inference of previous adapted model $f_\theta ^{{c_m - 1}}$ to super-resolve the downsampled I-frame $I_{{k_m}}^{LR}$:

\begin{equation}
\label{eq:5}
I_{{k_m}}^{SR} = f_\theta ^{{c_m - 1}}(I_{{k_m}}^{LR})
\end{equation}

We can calculate the PSNR between $I_{{k_m}}^{SR}$ and ${I_{{k_m}}}$ in terms of all possible patches as illustrated by Fig. \ref{fig:cps}. The time of calculating the PSNR map for 1080P frames is 0.1 seconds. Therefore, it is time-consuming to produce PSNR maps for all frames. For instance, when dealing with a 45-second video, it usually takes around 135 seconds to generate PSNR maps for all frames. Instead, we localize the positions of ${r\%}$ (r=20) most challenging patches in ${I_{{k_m}}}$, and then extract the patches from frames between ${I_{{k_m}}}$ and ${I_{{k_{m + 1}}}}$ according to the coordinates. As the frames between two I-frames are temporally consistent, the localized positions at I-frame can also choose reasonable patches on the in-between frames. Since I-frame is very sparse within a video, the computational cost is negligible. For a 45-second video, the total time of position localization and patch extraction is 0.7 seconds. However, the results of EMT using challenging patches are the same as results using all frames to some extent. In this way, the training efforts of EMT focus on challenging patches, resulting in faster convergence.

\section{Experiments}
In this section, we conduct extensive experiments to show the advantages of our method. The experimental details are given in Sec. \ref{sec:setting}. We first present the comparison with baseline and codec standards in Sec. \ref{sec:main}, and then compare EMT with other neural video delivery methods in Sec. \ref{sec:comparison}. We also conduct comprehensive ablation study to evaluate the contribution of each component in Sec. \ref{sec:abla}. In order to show the generalization ability, we report results across different scaling factors and architectures in Sec. \ref{sec:gen}.

\subsection{Experimental Details}
\label{sec:setting}
For meta-learning, we conduct two gradient updates for each individual task in the inner loop. After updating on all sampled tasks, we conduct one outer gradient update on the SR model. We randomly select training patches with a resolution of $144 \times 144$ and set mini-batch size as $16*n$, where n is the number of sampled tasks. Particularly, we set n to 15 and each task consists of 50 frames. We set inner learning rate $\alpha=0.5e-5$ and outer learning rate $\beta=1e-3$. We adopt Adam optimizer with $\beta_{1}=0.9$, $\beta_{2}=0.999$, $\epsilon=10^{-8}$. ESPCN serves as the default architecture, x2 is utilized as the default scaling factor and PSNR is the default metric. 

For fine-tuning, we conduct experiments on two video lengths, including 45 seconds and 2 minutes. The batch size is 16 and learning rate is $1e-4$. We set ${p_1}\%$ as 20\% and ${p_2}\%$ as 1\% to compress the parameters as default. We design three settings of our method, including S, M, and L. S and M settings adopt 0.1 epoch and 3 epochs for fine-tuning respectively. As for L setting, we alter ${p_1}\%$ to $100\%$.  We conduct all the experiments on NVIDIA V100 GPUs.

\subsection{Comparison with Baseline and Codec Standards}
\label{sec:main}

\begin{figure*}[t]
	\begin{center}
		\includegraphics[width=0.75\linewidth,height=7cm]{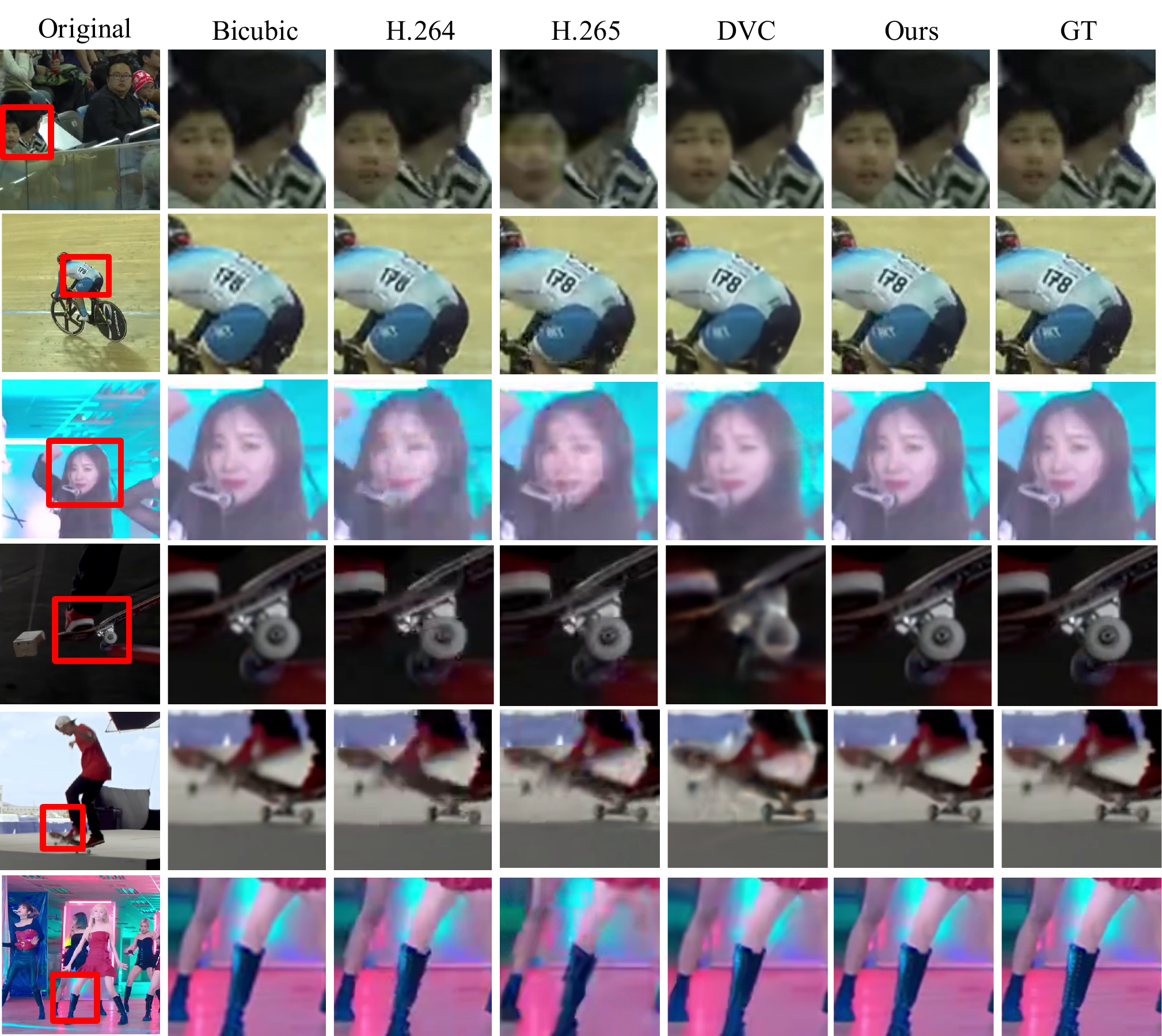}
	\end{center}
	\caption{Qualitative results. This figure shows the qualitative comparison against H.264, H.265 and DVC. Our method can restore better details compared with other methods. Best viewed by zooming x4.}
	\label{fig:vis}
\end{figure*}

In this section, we compare our method against the baseline \cite{yeo2018neural} and two codec standards. The baseline uniformly divides a video into chunks and trains one SR model for each chunk from scratch with 300 epochs. We denote the baseline as $C_{1-n}$. For the two commercial codec standards H.264 and H.265, we use ffmpeg with libx264 codec and libx265 codec to compress the HR videos to lower bit-rate while maintaining the resolution. The compressed videos are of the same storage size as our method (LR videos and SR models). We report three variants of our method under S, M, and L settings. Under the L setting, we aim to show the potential of meta-tuning by updating all the parameters for the first chunk. As shown in Tab. \ref{tab:main}, our method achieves better performance with less time and parameters compared with baseline \cite{yeo2018neural}. Our results with 0.1 epoch already outperform the baseline with 300 epochs. In terms of parameter compression, given a video with n chunks, we compress all models $n*P$ to $1*p_1\%P+(n-1)*p_2\%P$. In comparison with H.264 and H.265, our results outperform H.264 and H.265 in most cases as shown in Tab. \ref{tab:264265}. We also show the qualitative comparison in Fig. \ref{fig:vis}. As can be seen, our method can restore better details compared with codec standards.

\begin{table}[t]
	\begin{center}
	\caption{Comparisons with baseline \cite{yeo2018neural}. We show the results of our method under S, M, and L settings. Paras indicates the model parameters and P denotes the parameters of ESPCN. m and h in Time column represent minutes and hours respectively.}
	    \setlength{\tabcolsep}{1.3mm}{
		\begin{tabular}{llll|lll|llll}
			\hline
			\multirow{2}{*}{\textbf{}}&\multicolumn{3}{c}{\textbf {inter-45s}}& 
			\multicolumn{3}{c}{\textbf {sport-45s}} &
			\multicolumn{3}{c}{\textbf {game-45s}}\\
			
			& PSNR&Paras& Time&PSNR&Paras&Time& PSNR&Paras& Time\\\cmidrule{2-10}
			
			$C_{1-n}$        & 38.95 &9P        & 11.2h&46.03  &9P    &11.2h& 35.61  &9P   &11.2h    \\
			
			Ours(S)  & \textbf{39.08}   & 0.28P   &  1.2m & \textbf{46.11}   & 0.28P    & 1.2m& \textbf{36.32}     & 0.28P &  1.2m   \\
			Ours(M)  & 39.18   & 0.28P   & 7.6m& 46.25 & 0.28P    & 7.6m& 36.51     & 0.28P &  7.6m  \\
			Ours(L)    & 39.56   & 1.08P   & 55.5m & 46.41 & 1.08P    &1.76h& 37.09     & 1.08P &1.64h  \\
			\hline\hline
			\multirow{2}{*}{\textbf{}}&\multicolumn{3}{c}{\textbf {dance-45s}}& 
			\multicolumn{3}{c}{\textbf {vlog-45s}} &
			\multicolumn{3}{c}{\textbf {game-2min}}\\
			& PSNR&Paras& Time&PSNR&Paras& Time&PSNR&Paras& Time\\\cmidrule{2-10}
			$C_{1-n}$     & 43.47   &9P &11.2h  & 46.20  &9P&11.2h&34.46&24P&11.2h\\
			Ours(S)  & \textbf{44.13}    & 0.28P  & 1.2m & \textbf{46.48}   & 0.28P   & 1.2m &\textbf{34.35}&0.43P&1.2m \\
			Ours(M)  &44.24   & 0.28P   &7.6m& 46.71   & 0.28P&7.6m&34.50&0.43P&7.6m\\
			Ours(L)  &44.59  & 1.08P    & 1.53h&46.97  & 1.08P &48.6m&35.33 &1.23P&56.3m\\
			\hline
		\end{tabular}}
		\label{tab:main}
	\end{center}
\end{table}

\begin{table}[t]
	\centering
	\caption{Comparisons with H.264 and H.265. Under the same storage size, our PSNR results outperform H.264 and H.265 in most cases.}
	\begin{tabular}{lll|ll|ll|ll|ll}
		\hline
		\multirow{2}{*}{\textbf{}}&\multicolumn{2}{c}{\textbf {inter}}& 
		\multicolumn{2}{c}{\textbf {sport}} &
		\multicolumn{2}{c}{\textbf {game}}&
		\multicolumn{2}{c}{\textbf {dance}}&
		\multicolumn{2}{c}{\textbf {vlog}}\\
		
		& 45s&2min& 45s&2min& 45s&2min& 45s&2min& 45s&2min\\\cmidrule{2-11}
		
		H.264        & 36.32 &44.82        & 38.29  &36.75    & 37.29  &35.13   & 28.86   &29.36   & 42.37  &43.02    \\
		
		H.265  & 36.78   & 45.47   & 39.98   & 38.44    &\textbf{38.17}     & \textbf{36.87}& 30.83    & 31.06   & 43.35   & 43.97   \\
		Ours(M)  &\textbf{39.18}   & \textbf{46.73}   & \textbf{46.25}  & \textbf{40.44}    & 36.51     & 34.50& \textbf{44.24}   & \textbf{42.74}   & \textbf{46.71} & \textbf{48.66}  \\
		Storage(mB)    & 13.19   &  35.21  & 13.65  & 37.43    & 13.97     & 35.17& 13.81  & 36.80    & 13.83  & 36.37   \\
		\hline
	\end{tabular}
	\label{tab:264265}
\end{table}


\subsection{Comparison with Neural Video Delivery Methods}
\label{sec:comparison}

\begin{figure*}[t]
	\begin{center}
		\includegraphics[width=0.6\linewidth,height=5cm]{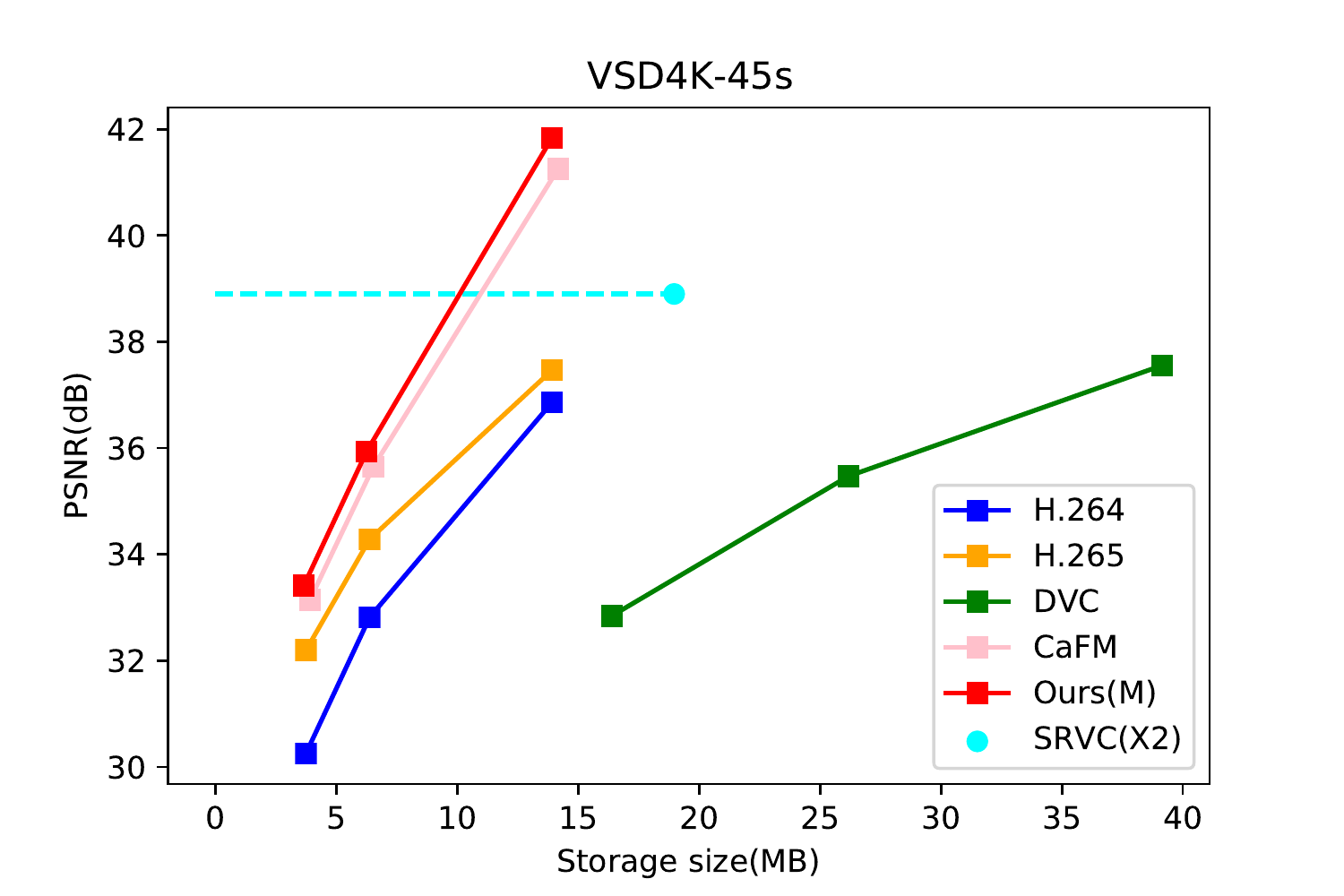}
	\end{center}
	\caption{Comparisons with neural video delivery methods in terms of PSNR and storage.}
	\label{fig:comp}
\end{figure*}

In this section, we compare our work with other neural video delivery methods. CaFM \cite{liu2021overfitting} uses content-aware models to compress parameters and achieve competitive performance. However, its joint training strategy takes huge computational cost. Therefore, CaFM is not practical for delivering long videos. SRVC \cite{khani2021efficient} also sequentially delivers the content-aware SR models by fine-tuning previous model. However, they fail to generalize on various architectures and have to train from scratch for a new video. Deep Video Compression (DVC) \cite{lu2019dvc} is an end-to-end DNN-based video compression method. We also compare our method with (DVC) at four different bitrate-distortion trade-off operating points $\lambda \in \{256, 512, 1024, 2048\}$ (DVC1, DVC2, DVC3, DVC4). In Fig. \ref{fig:comp} and Tab. \ref{tab:comp}, we demonstrate the advantages of our method in terms of accuracy, training time, and storage. In Fig. \ref{fig:comp}, we calculate the average PSNR and storage cost on 45s videos from VSD4K \cite{liu2021overfitting}. Under the same storage, our method outperforms other methods at most circumstances. Though CaFM achieves promising results, it takes a huge computational cost. As shown in Tab. \ref{tab:comp}, we demonstrate the trade-off between accuracy and training time. Our method shows competitive performance while maintaining low computational cost.

\begin{table}[t]
	\centering
	\caption{Comparisons with neural video delivery in terms of PSNR and training time. Red and blue indicate the best and the second best results among all methods.}
\setlength{\tabcolsep}{1.1mm}{
\begin{tabular}{c|cccccccccc}
			
			\hline
			\multirow{2}{*}{\textbf{}}&\multicolumn{2}{c}{\textbf {inter-45s}}&\multicolumn{2}{c}{\textbf {sport-45s}}&\multicolumn{2}{c}{\textbf {game-45s}}&\multicolumn{2}{c}{\textbf {dance-45s}}&\multicolumn{2}{c}{\textbf {vlog-45s}}\\\cmidrule{2-11}
			& Acc & Time& Acc & Time&  Acc & Time&  Acc & Time&  Acc & Time\\\hline
			C1-n     & \textcolor{blue}{38.95}     &  11.2h   & 46.03    &  11.2h & 35.61 &  11.2h& 43.47    &  11.2h & 46.20 &  11.2h \\\hline
			CaFM     & 38.90  &  10.2h&\textcolor{blue}{46.12}   &  10.2h&35.96  &10.2h&\textcolor{blue}{43.63}  &10.2h&\textcolor{blue}{46.45} &10.2h \\\hline
			SRVC     & 37.26  &  \textcolor{blue}{12.1m}& 41.38  &  \textcolor{blue}{12.1m}&33.34  &\textcolor{blue}{12.1m}&40.87  &\textcolor{blue}{12.1m}&45.59  &\textcolor{blue}{12.1m} \\\hline
			DVC1     &   31.98      & 35.6m&  35.52    &33.6m& 31.76   & 35.8m& 27.67   & 34.8m& 37.86   & 35.3m \\\hline
			DVC2     &   34.44     &   36.1m&  37.45    & 34.3m& 33.93  &36.0m& 32.46   & 35.2m& 39.92   & 35.8m\\\hline
			DVC3     &   36.60     & 37.1m&  39.58     & 34.8m& 36.10&36.5m& 34.40   & 35.4m& 41.67   & 36.2m\\\hline
			DVC4      &   38.70     & 38.0m&   41.28    &34.8m& \textcolor{red}{38.10}  &36.2m& 36.33   & 35.8m& 43.22   & 36.4m\\\hline
			Ours(M) & \textcolor{red}{39.18}     & \textcolor{red}{7.6m}& \textcolor{red}{46.25}    & \textcolor{red}{7.6m}&\textcolor{blue}{36.51} & \textcolor{red}{7.6m}&\textcolor{red}{44.24} & \textcolor{red}{7.6m}&\textcolor{red}{46.71} & \textcolor{red}{7.6m} \\\hline
		\end{tabular}}
\label{tab:comp}
\end{table}

\subsection{Ablation Study}
\label{sec:abla}

\textbf{Variants of EMT} We intend to evaluate the contribution of each component in EMT. Firstly, we compare our method with pretrained initialization, which is denoted as $P_{1-n}$. Similar to our meta-learned initialization, $P_{1-n}$ first initializes the SR model on DIV2K, and is normally finetuned on the meta-learning dataset $D_{N}$. We replace the meta-learned model of EMT by the normally 
finetuned model to obtain the results of $P_{1-n}$. To be mentioned, $P_{1-n}$ still utilizes gradient masking and challenging patch sampling for a fair comparison. Therefore, we are able to evaluate the effectiveness of meta-learned initialization. To evaluate the effectiveness of Challenging Patch Sampling (CPS), we remove the step of patch sampling in our full pipeline and the results are denoted as $MT_{1-n}$. As shown in Tab. \ref{tab:pre}, in order to achieve the same PSNR, $P_{1-n}$ takes extra cost compared with our meta-learned initialization. Meanwhile, our method outperforms $MT_{1-n}$ in regard to efficiency, demonstrating the effectiveness of CPS.

\begin{table}[!htb]
      \begin{center}
        \caption{Variants of EMT. We show the results of our method under S setting. MT stands for meta-tuning strategy. m and h in Time column represent minutes and hours respectively. P denotes the parameters of SR architecture (ESPCN).}
        \setlength{\tabcolsep}{2.0mm}{
       	\begin{tabular}{l|cc|llllll}
			\hline
			\multirow{2}{*}{\textbf{Method}}&\multirow{2}{*}{\textbf{MT}}&\multirow{2}{*}{\textbf{CPS}}&\multicolumn{3}{c}{\textbf {inter-45s}}& 
		
			\multicolumn{3}{c}{\textbf {sport-45s}}\\

			& & & PSNR&Paras& Time&PSNR&Paras&Time\\\cmidrule{1-9}
			$C_{1-n}$ & -&-&38.95 &9P&11.2h& 46.03  &9P &11.2h  \\
			
			$P_{1-n}$ & - & \checkmark & 39.08 &0.28P  &18.4m& 46.11 &0.28P &4.2m\\
			$MT_{1-n}$ & \checkmark& - &39.08&0.28P&\textcolor{blue}{9.7m}& 46.11&0.28P&\textcolor{blue}{2.7m}\\
			Ours(S) & \checkmark &\checkmark & 39.08 &0.28P  & \textcolor{red}{1.2m}&46.11 &0.28P&\textcolor{red}{1.2m} \\ \hline  
		\end{tabular}}
      \label{tab:pre}
      \end{center}
\end{table}

\textbf{Variants of Meta Learning} During meta-learning, we randomly sample 15 tasks and each task contains 50 frames. We also study the effect of different numbers of tasks and frames. For the number of tasks, we compare the results of 10, 15 and 20. For the number of frames, we evaluate the results of 10, 50 and 150. As shown in Tab. \ref{tab:maml}, all the variants achieve similar results, and the setting of 15 tasks with 50 frames already achieves competitive performance.

\begin{table}[!htb]
\begin{center}
\caption{Variants of meta-learning. Ours(S) adopt S setting.}
\setlength{\tabcolsep}{2.2mm}{
\begin{tabular}{l|ll|l|l|l|l}
\hline
& task & frame & inter-45s & sport-45s & game-45s & vlog-45s\\\hline
$C_{1-n}$       & -    & -  & 38.95   & 46.03 & 35.61 & 46.20   \\\hline
Ours(S) & 10   & 50    &   39.06   & 46.06 & 36.25 & 46.51    \\
Ours(S) & 15   & 50    &  39.08    & 46.11 & 36.32 & 46.48    \\
Ours(S) & 20   & 50    &  39.07    & 46.09 & 36.22 & 46.45   \\\hline\hline
Ours(S) & 15   & 10    &  39.09     & 45.94 & 36.34 & 46.57    \\
Ours(S) & 15   & 50    &  39.08    & 46.11 & 36.32 & 46.48     \\
Ours(S) & 15   & 150   &  39.03    & 46.20 & 36.39 & 46.58  \\\hline
\end{tabular}}
\label{tab:maml}
\end{center}
\end{table}

\textbf{Variants of Gradient Masking}
We conduct extensive experiments to explore the performance under different variants of gradient masking. Since our method uses different portions of parameters for fine-tuning the first chunk and other chunks. We report the results of ${p_1} \in \left\{ {10,20,30,100} \right\}$ and ${p_2} \in \left\{ {0.5,1,5,10} \right\}$. As can be seen from Tab.\ref{tab:grad}, we empirically set ${p_1} = 20$ and ${p_2} = 1$ as our default setting since it can already achieve satisfying results.

\begin{table}[!htb]
\begin{center}
\caption{Variants of gradient masking. Ours(S) adopt S setting.}
\setlength{\tabcolsep}{2.2mm}{
\begin{tabular}{l|ll|l|l|l|l}
\hline
& p1\% & p2 \% & inter-45s & sport-45s & game-45s & vlog-45s \\\hline
C1-n       & -       & -           & 38.95     & 46.03  & 35.61 & 46.20    \\\hline
Ours(S) & 10      & 1           & 39.04     & 46.00   & 36.28 & 46.42  \\
Ours(S) & 20      & 1           & 39.08     & 46.11  & 36.32 & 46.48  \\
Ours(S) & 30      & 1           & 39.11     & 46.13  & 36.33 & 46.49 \\
Ours(S) & 100     & 1           & 39.19     & 46.13   & 36.35 & 46.56  \\\hline
Ours(S) & 20      & 0.5         & 39.08     & 46.10   & 36.31 & 46.47 \\
Ours(S) & 20      & 1           & 39.08     & 46.11  & 36.32 & 46.48  \\
Ours(S) & 20      & 5           & 39.09     & 46.18   & 36.34 & 46.53 \\
Ours(S) & 20      & 10          & 39.11     & 46.23  & 36.37 & 46.57 \\\hline
\end{tabular}}
\label{tab:grad}
\end{center}
\end{table}

\subsection{The Generalization of Our Method}
\label{sec:gen}

\textbf{Generalization of Various Scaling Factors} We evaluate the generalization ability of EMT using ESPCN as the backbone across scaling factors x2, x3 and x4. As shown in Tab. \ref{tab:scale}, EMT outperforms $C_{1-n}$ under various scaling factors, demonstrating the generalization ability of EMT across various scaling factors.

\begin{table*}[!htb]
	\begin{center}
		\caption{Results of ESPCN on various scaling factors. We show the results of our method under M setting for fine-tuning.}
		\setlength{\tabcolsep}{1.6mm}{
		\begin{tabular}{cccccccccc}
			\hline
			\multirow{2}{*}{\textbf{}}&\multicolumn{3}{c}{\textbf {inter-45s}}& 
			\multicolumn{3}{c}{\textbf {sport-45s}} &
			\multicolumn{3}{c}{\textbf {vlog-45s}}\\
			
			& x2 &x3&x4&x2&x3&x4& x2&x3&x4\\\cmidrule{2-10}
			
			$C_{1-n}$     & 38.95 & 32.19     & 28.73& 46.03 &40.43  & 37.21 & 46.20  & 41.68   &39.52 \\\hline
			Ours(M) & 39.18 & 32.55   & 29.05 & 46.25 & 40.51  & 37.22  & 46.71 & 42.25  & 40.08\\
			
			\hline\hline
			\multirow{2}{*}{\textbf{}}&\multicolumn{3}{c}{\textbf {dance-45s}}& 
			\multicolumn{3}{c}{\textbf {game-45s}} &
			\multicolumn{3}{c}{\textbf {city-45s}}\\
			
			& x2 &x3&x4&x2&x3&x4& x2&x3&x4\\\cmidrule{2-10}
			$C_{1-n}$     & 43.47 & 36.86 & 35.22 & 35.61 &30.67  & 28.80 &   36.44   & 31.60   &   29.23  \\\hline
			Ours(M) & 44.24 &37.42& 35.88& 36.51 & 31.13  & 29.01&  36.42  &  31.56   &  29.16 \\\hline
		\end{tabular}}
		\label{tab:scale}
	\end{center}
\end{table*}

\textbf{Generalization of Various Efficient Backbones}
In this part, we present the results of our method using different efficient SR backbones. We evaluate our method on 45 seconds videos and adopt three additional efficient backbones, including SRCNN \cite{dong2014learning}, FSRCNN \cite{dong2014learning}, and EDSR with one residual block in body (EDSR-1). As shown in Tab. \ref{tab:gen}, our method also generalizes well to different efficient backbones, validating the generalization ability of our method.

\begin{table*}[!htb]
	\begin{center}
	\caption{Results of various SR architectures. We show the results of our method under M setting for fine-tuning.}
	    \setlength{\tabcolsep}{1.2mm}{
		\begin{tabular}{ccccccccccc}
			\hline
			\multirow{2}{*}{\textbf{Backbone}}&\multicolumn{2}{c}{\textbf {inter-45s}}& 
			\multicolumn{2}{c}{\textbf {sport-45s}} &
			\multicolumn{2}{c}{\textbf {game-45s}}&
			\multicolumn{2}{c}{\textbf {dance-45s}}&
			\multicolumn{2}{c}{\textbf {vlog-45s}}\\
			
			& $C_{1-n}$&Ours&$C_{1-n}$&Ours&$C_{1-n}$&Ours& $C_{1-n}$&Ours& $C_{1-n}$&Ours\\\cmidrule{2-11}
			
			SRCNN        & 39.02 & 39.06  &  46.21 &46.19  & 35.37 & 35.64 &  43.28 & 43.92 & 46.25 &  46.42  \\
			
			FSRCNN  & 39.06   & 39.25   & 46.29   & 46.32    & 35.84     & 35.90& 43.61    & 44.08   & 46.09   & 46.47   \\
			EDSR-1  & 39.17  & 39.13   & 45.99  & 46.02    & 35.51     & 35.58& 44.04   & 44.05   & 46.28   & 46.28  \\
			
			\hline
		\end{tabular}}
		\label{tab:gen}
	\end{center}
\end{table*}

\begin{table*}[!htb]
	\begin{center}
		\caption{Comparisons with H.264 and H.265 on long videos. We show the results of our method using 3 epochs for fine-tuning. The storage is measured in megabytes.}
		\setlength{\tabcolsep}{1.2mm}{
		\begin{tabular}{lll|ll|ll|ll}
			\hline
			\multirow{2}{*}{\textbf{}}&\multicolumn{2}{c}{\textbf {vlog-5min}}& 
			\multicolumn{2}{c}{\textbf {vlog-10min}} &
			\multicolumn{2}{c}{\textbf {vlog-20min}} &
			\multicolumn{2}{c}{\textbf {vlog-30min}}\\
			
			& PSNR&Storage&PSNR&Storage& PSNR&Storage& PSNR&Storage\\\cmidrule{1-9}
			
			
			Ours(M)  & 37.67  & 18.62   & 38.33  & 35.64  & 38.31     & 71.19 & 38.41    & 145.12\\
			
			H.264        & 34.68 &18.62   & 37.15  &35.64   & 35.29  &71.19 & 35.02     & 145.12 \\
			
			H.265  & 36.75 &18.62 & 35.78    & 35.64  & 37.18    & 71.19   & 37.07     & 145.12\\

			\hline
		\end{tabular}}
		\label{tab:264265long}
	\end{center}
\end{table*}
\section{Extension to Long Videos}

In this section, we extend EMT to long videos beyond 2 minutes. Since former works \cite{liu2021overfitting,yeo2018neural} take too much time to train the content-aware SR models, we only compare with commercial codec standards. Directly applying EMT to long videos may achieve degraded results since the temporal consistency between neighboring chunks is not always true for long videos. Therefore, in order to extend EMT to long videos, we first divide the video frames into groups and apply EMT to each group. To be specific, we extract all the I-frames from the input video and make each group contain 30 I-frames. As shown in Tab. \ref{tab:264265long}, our method outperforms commercial codec standards even on long videos, showing the great potential of our method.



\section{Conclusion}
To pave the way for practical applications, we propose Efficient Meta-Tuning (EMT) to significantly reduce the computational cost of neural video delivery. Instead of training from scratch, EMT sequentially fine-tunes a meta-learned model to deliver the content-aware SR models of the input video. Gradient masking is introduced to select partial parameters for fine-tuning, compressing all the models into one shared model and a few private parameters. In addition, we also present a sampling strategy to extract the challenging patches for fine-tuning, further reducing the cost of EMT. We conduct detailed comparisons with the commercial codec standards and other neural video delivery methods to demonstrate the advantages of our approach. We hope this paper can inspire future work on neural video delivery.

\clearpage
%
%
\bibliographystyle{splncs04}
\bibliography{egbib}
\end{document}